\documentclass{article}

\usepackage{arxiv}

\usepackage[utf8]{inputenc} 
\usepackage[T1]{fontenc}    
\usepackage{hyperref}       
\usepackage{url}            
\usepackage{booktabs}       
\usepackage{amsfonts}       
\usepackage{nicefrac}       
\usepackage{microtype}      
\usepackage{lipsum}		
\usepackage{graphicx}
\usepackage{natbib}
\usepackage{doi}

\usepackage{latexsym}
\usepackage{amsmath,amsfonts}
\usepackage{multirow}
\usepackage[noend]{algpseudocode}
\usepackage{algorithm}
\usepackage{verbatim}

\newcommand{\tit}[1]{\textit{#1}}

\title{Enriching language models with graph-based context information to better understand textual data}

\date{} 					

\author{ \href{https://orcid.org/0000-0001-5077-3480}{\includegraphics[scale=0.06]{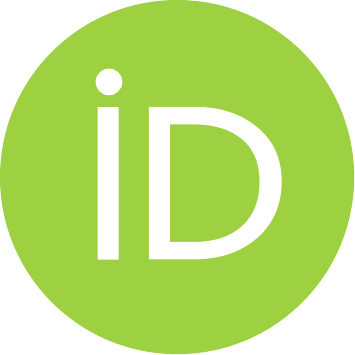}\hspace{1mm}Albert Roethel}
\\
	Faculty of Mathematics and Information Science\\
	Warsaw University of Technology\\
	Warsaw, Poland\\
	\texttt{roethel.albert@gmail.com} \\
	\And
	\href{https://orcid.org/0000-0001-7714-4844}{\includegraphics[scale=0.06]{orcid.pdf}\hspace{1mm}Maria Ganzha} \\
	Faculty of Mathematics and Information Science\\
	Warsaw University of Technology\\
	Warsaw, Poland\\
	\texttt{maria.ganzha@pw.edu.pl} \\
    \And
	\href{https://orcid.org/0000-0002-3407-7570 }{\includegraphics[scale=0.06]{orcid.pdf}\hspace{1mm}Anna Wróblewska} \\
	Faculty of Mathematics and Information Science\\
	Warsaw University of Technology, Warsaw, Poland\\
    weSub, Warsaw, Poland\\
	\texttt{anna.wroblewska1@pw.edu.pl} \\
}

\renewcommand{\shorttitle}{Language models with graph-based context information}

\hypersetup{
pdftitle={Enriching language models with graph-based context information to better understand textual data},
pdfsubject={cs.LG, cs.CL, cs.NE, cs.AI},
pdfauthor={Albert Roethel, Maria Ganzha, Anna Wróblewska},
pdfkeywords={Natural language processing, Graph-based machine learning, Representation learning, Graph neural networks, Transformers, Text classification},
}

\begin{document}
\maketitle

\begin{abstract}
A considerable number of texts encountered daily are somehow connected with each other. For example, Wikipedia articles refer to other articles via hyperlinks, scientific papers relate to others via citations or (co)authors, while tweets relate via users that follow each other or reshare content. Hence, a graph-like structure can represent existing connections and be seen as capturing the ``context'' of the texts. The question thus arises if extracting and integrating such context information into a language model might help facilitate a better automated understanding of the text. In this study, we experimentally demonstrate that incorporating graph-based contextualization into BERT model enhances its performance on an example of a classification task. Specifically, on Pubmed dataset, we observed a reduction in error from 8.51\% to 7.96\%, while increasing the number of parameters just by 1.6\%.
\\
\\
Our source code:~\url{github.com/tryptofanik/gc-bert}

\end{abstract}


\section{Introduction}

The abundance of texts and data on the Internet makes it challenging for individuals to navigate and understand the information they encounter. From social media platforms, such as Twitter, LinkedIn, and Facebook, to scientific papers, the quality and usefulness of existing texts can vary greatly. Therefore, an automatic tool is needed to process and utilize this vast amount of information effectively, to reduce information overload and select the most appropriate information for the user.

One possible application of such a tool could be effective text classification, which could deliver better text recommendations, hate speech and/or misinformation detection, and play a role in text translation and/or summarization, among others. For example, the authors of~\cite{hateful_users_on_twitter} show that Twitter haters tend to interact intensely with each other, forming a clustered network. This finding highlights also the importance of analyzing the graph-represented context of social media posts when processing them.

Separately, authors in scientific research often refer to other articles to build upon existing knowledge and compare their work to the state-of-the-art. These references can be captured and aggregated, for instance, through a graph neural network, to provide more context to a language model in NLP tasks. Similarly, websites such as Wikipedia, where articles refer to each other, can benefit from similar contextualization of knowledge. For instance, the language model can be trained to include the context of a given article for text classification tasks. Thus, it can be conjectured that incorporating graph-represented context information into language models can enhance the understanding of processed texts. This, in turn, can result in more accurate and efficient models.


In this context, the aim of this contribution is to enhance a deep-learning language model by incorporating graph-represented information. Such information captures existing links between documents and is referred to as a graph context. Graph context becomes one of the inputs into a deep learning model, consisting of two components: the GNN (Graph Neural Network) and the LM (Language Model). As a result, after exploring various ways of connecting the two components, we also introduced a new deep learning model named GCBERT (Graph Context BERT). GCBERT slightly outperforms the BERT model in the text classification task on Pubmed dataset.\footnote{Our source code and details concerning tuning and setting hyperparameters and seeds are available online on~\url{github.com/tryptofanik/gc-bert}.
} 

The remaining parts of this contribution are organized as follows. Section~\ref{sec:related-work} highlights related work. Section~\ref{sec:methodology} describes the proposed approach and the datasets used in experiments. Following, in Section~\ref{sec:results}, experimental results are presented and analysed. Future research directions, including the potential for graph-based networks, are discussed in the last section.

\section{Related Work}\label{sec:related-work}

So far, several works have explored different ways of combining graph neural networks with language models, such as BERT (Bidirectional Encoder Representations from Transformers)~\cite{bert}. 
The following section will discuss how GNN and transformer models have been used to solve various Natural Language Processing (NLP) tasks. 

In~\cite{jeong_context-aware_2020}, the BERT-GCN architecture is proposed to recommend context-aware paper citations. The BERT model's input is a text that includes [REF] tokens, which are used to inform the model where the reference is needed. BERT creates a vector representation of the document to which the user seeks references. In parallel, the citation graph representation is created using a Variational Graph AutoEncoder with a set of papers from a selected domain and their reference network. Both representations are concatenated and processed by a feed-forward neural network (FF-NN), which produces a softmax output indicating the citation label. This model aggregates the entire graph into one vector for FF-NN  and does not utilize graph information within the BERT model.


In~\cite{OstendorffBBSRG19}, BERT is combined with metadata and graph-based author data with the goal of text classification. The authors use the PyTorch BigGraph embeddings~\cite{MLSYS2019_e2c420d9} to create authors' representations concatenated with metadata and fed to an FF-NN to make predictions. Within this architecture, the language model also lacks awareness of contextual data in the analyzed document.

In contrast to the parallel deep learning architectures that analyze the data independently,  compositional models have been proposed. Here, data $x$ is taken by the model $f$, and the output is consumed by the model $g$, which can be written as $g(f(x)) = (g \circ f) (x)$. The benefit of this solution -- pipeline of models -- is that it can inject what was learned and noticed by one model into the other in the pipeline. Among these approaches, BertGCN~\cite{lin-etal-2021-bertgcn} is a straightforward implementation of the composition of BERT and GCN (graph convolutional networks). It extracts text-based representations from each document using BERT, which is then used as input in the GCN. The heterogeneous graph structure, constructed similarly as in the TextGCN study~\cite{Yao_Mao_Luo_2019}, utilizes documents as nodes, and additional nodes were created based on words and their concurrence in the documents. However, BertGCN does not consider any interconnections between texts other than semantic meaning.
Separately, authors of~\cite{gao_gating_2021} combined GCN and BERT, using a gating mechanism, to create a compositional model. A heterogeneous text graph is created and then processed by GCN to produce a set of hidden states of documents. In parallel, BERT embeddings are derived and combined with the hidden states using the gating mechanism. The result is then processed by another GCN network, allowing predictions to be retrieved from tokens representing particular documents.

The short-text graph convolutional network
~\cite{stgcn} is another model used for text classification. Documents are connected with words they contain and topics they refer to. GCN then processes the graph, while each node representation is calculated for triples consisting of documents, topics and  words. 
BERT is used to produce abstracted and contextualized tokens for each word in a document. These tokens are then concatenated with GCN-derived word node representations and used as input for a Bi-LSTM (Bidirectional Long Short-Term Memory) network. The final output category of the document is produced by combining the document representation 
and the LSTM vector.
%

Another group of approaches is based on Knowledge Graphs (KG), which represent real-world entities and their relationships as a graph structure. Nodes in the graph correspond to entities
, while edges correspond to the relationships between them
.
KG-BERT (Knowledge-Graph BERT), introduced in~\cite{kg-bert}, is a model that utilizes BERT to inject entities and relationships into the language model. Entities and relationships are represented as vectors and fed into the BERT model, with separation tokens applied to distinguish them. Unlike other graph-based models, KG-BERT does not utilize GNNs, instead relying on the language model to understand the language and the graph relationships. KG-BERT has been applied as a tool for knowledge graph completion.

An alternative approach to integrating knowledge graph information into a language model in a context-aware manner is presented in~\cite{Yu_Zhu_Yang_Zeng_2022}.  The JAKET architecture consists of a knowledge module that produces entity embeddings corresponding to concepts in KG and a language module that analyzes a text. The knowledge module uses a graph attention network to embed entities, while the language module comprises two language models, LM1 and LM2. LM1 is shared and operates on pure text data, providing embeddings for LM2. The knowledge graph embeddings are integrated with LM1's in LM2, allowing access to textual and contextual knowledge graph information. 
The prediction phase utilizes the output of LM2. JAKET 
provides semantic context, using knowledge graph-stored word embeddings. Compared to models created in our study (like GC-BERT), JAKET does not utilize graph information additional to the given text, e.g. relations between texts given by authorships or retweets.


The study presented in \cite{10.1007/978-3-031-01333-1_25} addresses the authorship verification problem using a combination of a language model and a graph neural network called LG4AV. The approach calculates a vector representation for each author based on their article embeddings, which is then treated as a classification token ([CLS-a]) in the BERT language model. The BERT processes the document and produces the updated classification token (LM($d$)), while the GNN computes the author vectors updated by the ``neighbours''. The final prediction $f(a, d)$ is obtained by aggregating the vectors multiplied by the LM($d$). LG4AV integrates graph context into the language model, potentially improving judgment for authorship verification. However, its limitations are that only the language model is trainable, while graph representations are static. Additionally, the design of the architecture enforces the concrete structure of the data, in which documents must be authored by a group of people. (These constraints of the model will be addressed in the current work.)

So far, many researchers have worked on linking NLP models with semantics constructed as knowledge graphs that allow expressing mapping, organizing, and relating ideas, entities, and concepts expressed in natural language~\cite{10.1145/3447772}. However, texts do not have to be connected just because they are related semantically. It is enough that they were created by the same user or share a common hashtag on the social media platform. It is hypothesized that some additional predictive information might be available in other related texts, and this could improve ``prediction'' results. The NLP community has not thoroughly researched this aspect of the text analysis. In the literature, the use of semantics in NLP models is limited primarily to understanding language and its nuances. Current research does not cover additional information hidden in how the texts are organized and related within real-world systems. We have noted that a limited number of studies in the literature currently analyze the network of documents in a way that leverages rich and diversified graph context information. Also, the existing architectures either cannot 
train simultaneously text and graph representations; each time the model is trained, only one representation is trained. Thus, both representations are not fully dynamic throughout the training process.

\section{Proposed Approach}\label{sec:methodology}

Our work goes beyond the classical semantic enhancement of the language models and seeks more predictive power in understanding how texts relate to each other that can be given by additional structural information --additional to the text semantics. The architectures developed in this work aim to be flexible enough for various scenarios where texts have relevant, meaningful, and logical connections and could be used for both graph and text-based downstream tasks. 

More formally, this study works on datasets with the following properties: (1)~a set of $n$ documents $D = \{d_1, d_2, d_3, ..., d_n\}$, (2)~a set of direct or indirect connections between documents, represented as a set of edges $E = \{(d_1, d_2), (d_1, d_5), ..., (d_i, d_j)\}$. The input must be a graph of interconnected documents. Noteworthy, the graph information should add some new information to the documents and thus have the potential to improve the final prediction results. Further, in our notation, an $n \times n$ adjacency matrix $A$ provides information about edges (i.e. interconnections between documents), such that $A[i,j] = 1$, if document $i$ relates to document $j$. 

This work is devoted to developing a deep-learning architecture consuming graph and text information that is able to train text representations and graph context representations simultaneously. It is also possible to utilize our architectures while having only access to static representations of texts or nodes. In that case, these representations are constant throughout the entire experiment and can not be changed during training. It can be achieved either by using some deterministic function to obtain these representations like TF-IDF (term frequency-inverse document frequency)~\cite{tfIdf-explanations} or by freezing weights of a deep learning model during the training. 


\subsection{GNN+LM Architectures}
%
%
The basic building blocks, used for this study, are the Graph Convolutional Network (GCN)~\cite{Kipf:2016tc} and the encoder of the BERT model~\cite{bert} (uncased, based version with 110M parameters\footnote{\url{https://huggingface.co/bert-base-uncased}}).

It was chosen due to its simplicity compared to different LMs and extensive usage in other studies. 
However, any other language model could be chosen for this task, as the only thing required is to encode the text into a representation vector. Also, any other GNN architecture can be utilized instead of GCN; the only requirement is that it should take as an input vector representation of nodes and an adjacency matrix.

As part of the performed research, we experimented with four ways to connect a language model and a graph neural network: \tit{late fusion}, \tit{early fusion -- GCBERT}, \tit{compositional architecture GNN(BERT)}, and \tit{looped GCBERT}. 

The processed text and node representations calculated by the LM and GNN are stored in matrices: $T$ (text representation) and $N$ (node representation), respectively, and information about interconnections between nodes is in the $A$ adjacency matrix.
%
%
The initialization of the $T$ and $N$ matrices is necessary for the LM+GNN model training. The $T$ matrix is initialized using the pretrained BERT model, while the $N$ matrix is filled with the GNN model, which utilizes different transformations depending on the following architectures and their training algorithms. 

In all GNN+LM experiments, the total model size is 111.84M parameters (110.07M BERT + 1.77M GCN).




\subsubsection{Late Fusion}

\begin{figure}[htp]
    \centering
    \includegraphics[width=0.6\textwidth]{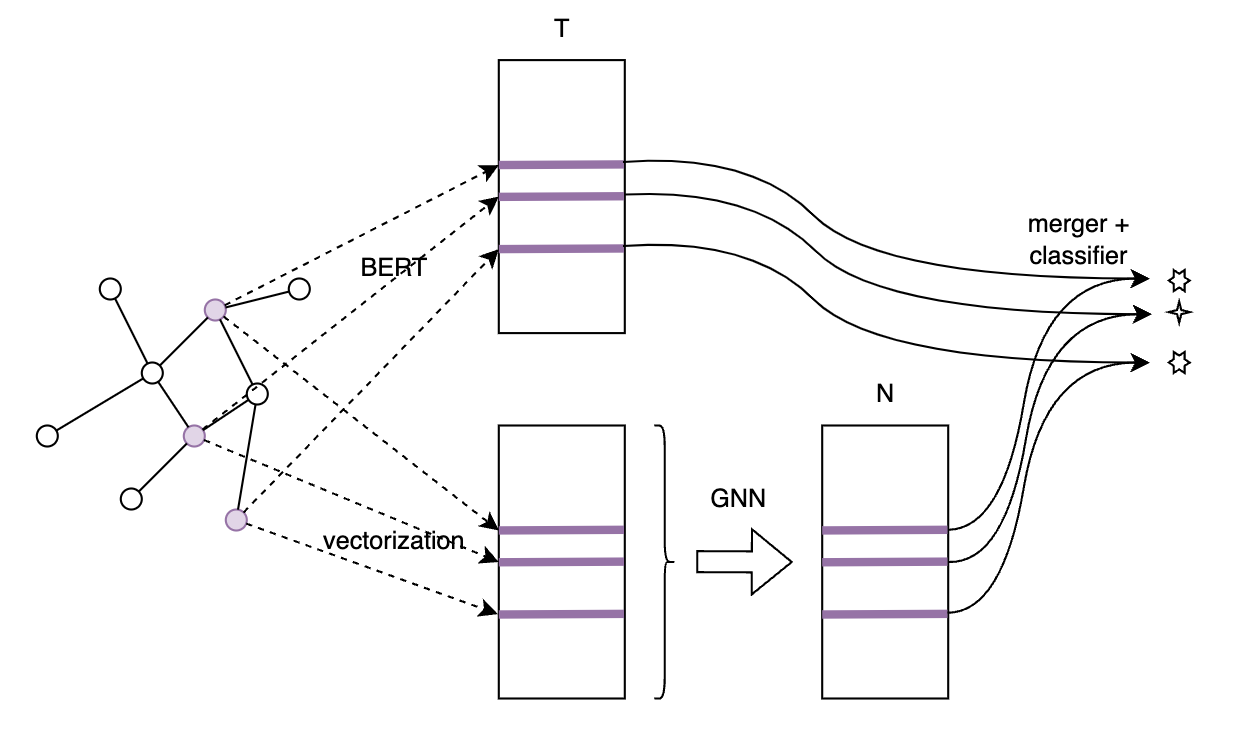}
    \caption{Basic \tit{Late fusion} architecture. Note: nodes in the graph are text documents to classify, edges are interconnections between documents (e.g. citations), other notations are given in Algorithm~\ref{alg:late_fusion}, the outputs are predictions of three classes. Apart from vectorized texts, GNN also consumes the adjacency matrix $A$, which was not shown in the figure for clarity.}
    \label{fig_late_fus_arch}
    \end{figure}

A parallel \tit{late fusion} architecture is the most basic combination of GNN and BERT (see  Figure~\ref{fig_late_fus_arch}). Here, BERT receives pure text and creates textual representation vectors. In parallel, each document (input text) is vectorized. It can be done using TF-IDF or reusing representations created by BERT (in our experiments we reused BERT representations). The vectorized texts and the graph data are sent to the GNN model, producing a node representation of each text. Next, node and text representations are merged and processed by a classifier (see Figure~\ref{fig_late_fus_arch} and Algorithm~\ref{alg:late_fusion}).

In this architecture, each model (GNN and LM) works on the data type it was designed for. However, they do not have a direct opportunity to ``strengthen each other''. One model cannot utilize the information contained in the input of the other. Here, BERT analyzes the text to extract semantics and context, but it cannot access the graph context. In contrast, GNN works on both textual and graph data, but the text data is limited as it is compressed into a single vector. The classifier is the first module that takes into account both sources of information.
As it turned out further, it might not have enough flexibility to extract and capture the information hidden in the merged input.

 \begin{algorithm}[h!]
    \caption{Basic \tit{Late fusion} algorithm. $D$ -- set of documents, $T$ -- text representations, $A$ -- adjacency matrix, $N$ -- node representations, $E$ -- number of epoches, $I$ -- iterator over training set, $i$ -- batch indices.}\label{alg:late_fusion}
    \begin{algorithmic}
    \State $T \gets \text{BERT}(D)$ \Comment{Initialize T and N}
    \State $V \gets \text{VECTORIZER}(D)$
    \State $N \gets \text{GNN}(V, A)$
    \For{$e \in \{1,2,3,...,E\}$} \Comment{Iterate over epoches}
        \For{$i \in I$} \Comment{Iterate over batches}
            \State $T[i] \gets \text{BERT}(D[i])$
            \State $N[i] \gets \text{GNN}(V[i], A)$
            \State $x \gets \text{MERGER}(T[i], N[i])$
            \State $\hat{y} \gets \text{CLASSIFIER}(x)$
            \State $err \gets \text{LOSS}(\hat{y}, y)$
            \State Backprop
        \EndFor
    \EndFor
    \end{algorithmic}
    \end{algorithm}

\subsubsection{Early Fusion -- GCBERT}
\begin{figure}[htp]
\centering
\includegraphics[width=0.6\textwidth]{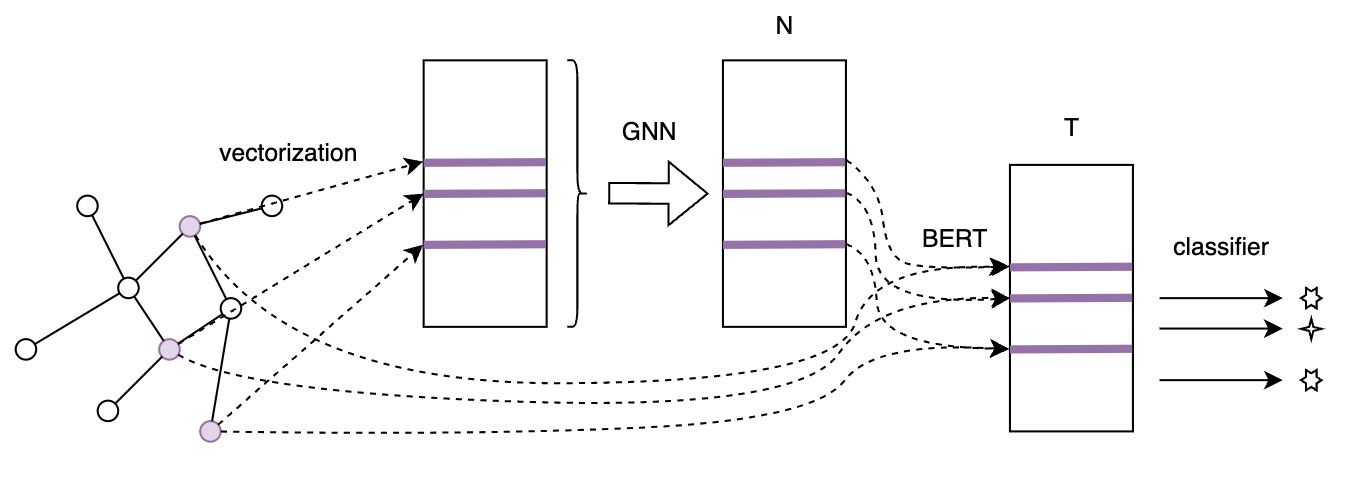}
\caption{Basic \tit{Early fusion} -- GCBERT architecture. (Notation is the same as in Figure~\ref{fig_late_fus_arch})}
\label{fig_early_arch}
\end{figure}

%
The second architecture is \tit{early fusion}, where texts are vectorized into static vectors (via TF-IDF or finetuned  and then frozen BERT)
processed by GNN, which creates node representations. They are then inserted into the input token sequence of the text and processed by trainable BERT, which subsequently produces the text representations. Next, a classifier uses those to make a prediction -- see Figure~\ref{fig_early_arch} and Algorithm \ref{alg:early_fusion}. 
In this scenario, GNN still has access only to the graph and to the static textual data, while BERT is now fed directly with both textual and graph information. It is worth emphasizing that backpropagation will flow from BERT to GNN and can enforce some adjustments in the GNN model.
To highlight the fact that BERT is also analysing the graph context of each of the documents, it is now referred to as GCBERT (\textbf{G}raph \textbf{C}ontext \textbf{BERT}).
%
%

 \begin{algorithm}
    \caption{Basic \tit{Early fusion} GCBERT algorithm. $D$ -- set of documents, $T$ -- text representations, $A$ -- adjacency matrix, $N$ -- node representations, $E$ -- number of epoches, $I$ -- iterator over training set, $i$ -- batch indices, GCBERT -- BERT that consumes graph context data.}\label{alg:early_fusion}
    \begin{algorithmic}
    \State $T \gets \text{BERT}(D)$ \Comment{Initialize T and N}
    \State $V \gets \text{VECTORIZER}(D)$
    \State $N \gets \text{GNN}(V, A)$
    \For{$e \in \{1,2,3,...,E\}$} \Comment{Iterate over epoches}
        \For{$i \in I$} \Comment{Iterate over batches}
            \State $N[i] \gets \text{GNN}(V[i], A)$
            \State $T[i] \gets \text{GCBERT}(D[i], N[i])$
            \State $\hat{y} \gets \text{CLASSIFIER}(T[i])$
            \State $err \gets \text{LOSS}(\hat{y}, y)$
            \State Backprop
        \EndFor
    \EndFor
    \end{algorithmic}
    \end{algorithm}

\subsubsection{Compositional Architecture -- GNN(BERT)} %
\begin{figure}[htp]
    \centering
    \includegraphics[width=0.6\textwidth]{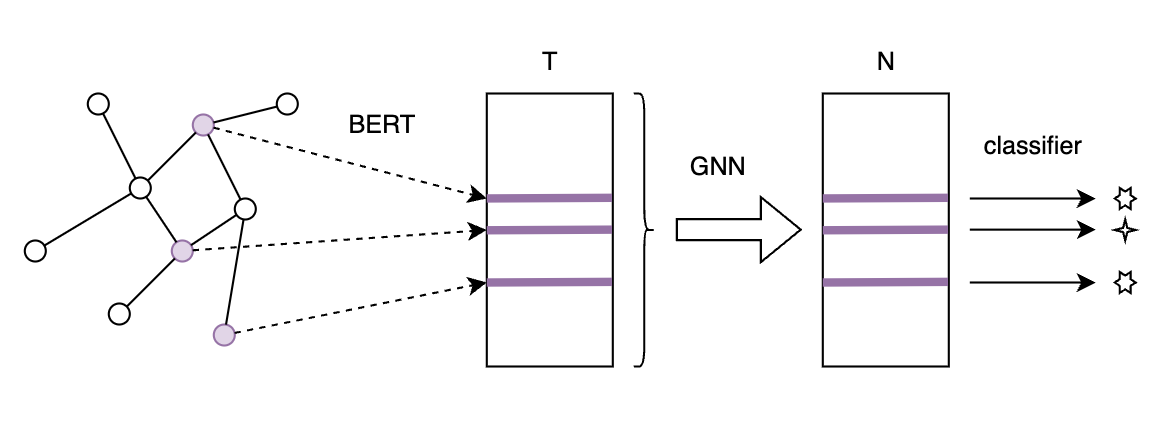}
    \caption{Basic \tit{Compositional architecture} GNN(BERT). (Notation is the same as in Figure~\ref{fig_late_fus_arch})}
    \label{fig_compositional_arch}
    \end{figure}

The third architecture is \tit{compositional architecture} GNN(BERT), where BERT processes the pure text data to create text representations, which is then consumed by GNN to produce node representations for classification -- see Figure~\ref{fig_compositional_arch} and Algorithm \ref{alg:compositional}. 
In this scenario, GNN is supplied with dynamic BERT textual representation that should contain more information than vectors obtained via TF-IDF or static, precalculated BERT vectors. However, the transformer does not have direct access to the information about the graph context of the processed document.
Similarly, as in the previous architecture -- GCBERT, it is worth emphasizing that in this architecture, the backpropagation will flow from GNN to BERT and can enforce some adjustments in the BERT model.

    \begin{algorithm}
    \caption{Basic \tit{Compositional architecture} GNN(BERT) algorithm. $D$ -- set of documents, $T$ -- text representations, $A$ -- adjacency matrix, $N$ -- node representations, $E$ -- number of epoches, $I$ -- iterator over training set, $i$ -- batch indices.}\label{alg:compositional}
    \begin{algorithmic}
    \State $T \gets \text{BERT}(D)$ \Comment{Initialize T and N}
    \State $N \gets \text{GNN}(T, A)$
    \For{$e \in \{1,2,3,...,E\}$} \Comment{Iterate over epoches}
        \For{$i \in I$} \Comment{Iterate over batches}
            \State $T[i] \gets \text{BERT}(D[i])$
            \State $N[i] \gets \text{GNN}(T[i], A)$
            \State $\hat{y} \gets \text{CLASSIFIER}(N[i])$
            \State $err \gets \text{LOSS}(\hat{y}, y)$
            \State Backprop
        \EndFor
    \EndFor
    \end{algorithmic}
    \end{algorithm}
    
\subsubsection{Looped GCBERT}
\begin{figure}[htbp]
    \centering
    \includegraphics[width=0.6\textwidth]{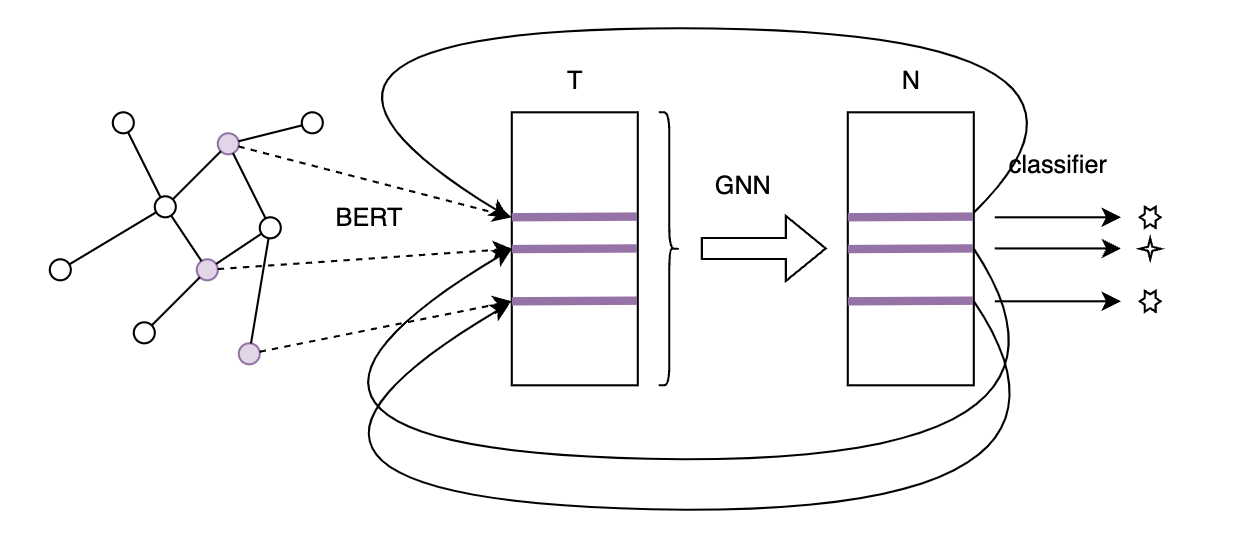}
    \caption{Basic \tit{Looped GCBERT} with BERT vectorization. (Notation is the same as in Figure~\ref{fig_late_fus_arch})}
    \label{fig_looped_early_arch}
    \end{figure}

The fourth architecture is \tit{Looped GCBERT}, which combines \tit{early fusion} and \tit{compositional architecture}. 	
BERT receives textual data and the latest node context vector and produces the graph-augmented text representation. 	
It is then processed by GNN, which produces updated node representations then used for classification -- see Figure~\ref{fig_looped_early_arch} and Algorithm~\ref{alg:looped}.	
During each epoch of training, the text representation of a document is produced based on the node representation from the previous epoch, allowing both components to interchange what they have learned directly.	
This model is a combination of early fusion and compositional architecture in which the whole architecture works on dynamic (trainable) text and node representations. It means that GNN and BERT must be unfrozen in this architecture choice. 
%
%

\begin{algorithm}[]
    \caption{Basic \tit{Looped GCBERT} architecture algorithm. $D$ -- set of documents, $T$ -- text representations, $A$ -- adjacency matrix, $N$ -- node representations, $E$ -- number of epoches, $I$ -- iterator over training set, $i$ -- batch indices.}\label{alg:looped}
    \begin{algorithmic}
    \State $T \gets \text{BERT}(D)$ \Comment{Initialize T and N} \State $N \gets \text{GNN}(T, A)$
    \For{$e \in \{1,2,3,...,E\}$} \Comment{Iterate over epoches}
        \For{$i \in I$} \Comment{Iterate over batches}
            \State $T[i] \gets \text{GCBERT}(D[i], N[i])$ 
            \State $N[i] \gets \text{GNN}(T[i], A)$
            \State $\hat{y} \gets \text{CLASSIFIER}(N[i])$
            \State $err \gets \text{LOSS}(\hat{y}, y)$
            \State Backprop
        \EndFor
    \EndFor
    \end{algorithmic}
    \end{algorithm}
    
\subsubsection{Architectural Modifications}
%
The modifications that can be applied to our models include:
\begin{itemize}
\item \tit{Skip connections}: transferring node representation $N$ or text representation $T$ into the classifier (the last element of all the architectures)  to prevent information loss and allow later modules to analyze all the available representations. Before the classifier, both the  representations should be merged with a merger function (as it is in a basic \tit{late fusion} algorithm). In this case, the merger has both representations at its disposal, regardless of whether one was created using the other. 
\item \tit{Freezing/Unfreezing}: models can be frozen to obtain static representations or unfrozen to obtain dynamic representations, which can potentially improve performance but are more challenging to train. In our experiments, we mostly unfroze representations so that the models GNN or BERT are trainable throughout the training process. Otherwise, if we froze them, we also stated it clearly in the description of a particular test.
\item \tit{Merger} function: the merging of $T[i]$ and $N[i]$ vectors before classifier (as it is in a basic \tit{late fusion} algorithm) can be concatenated, added in an element-wise manner, or combined using other functions like max, to preserve information and reduce parameters.
\end{itemize}

\subsubsection{Language Model Augmentation with Graph Context Token}
%
%
The standard BERT architecture takes as input a sequence of tokens, each of which can represent words, subwords, or special signs like '?', '-', etc. BERT also uses special tokens, such as the classification token $\text{[CLS]}$, used as a document embedding after the final BERT layer. Each token is represented by a vector in $\mathbb{R}^{n}$ space, and a positional embedding is added to each vector to convey information about the token's order in the sequence~\cite{bert}.
%

The way we insert a graph context token into the language model is essential for its proper utilization and for boosting the model's performance. 
In this study, when the BERT architecture is augmented with the output of GNN, it is carried out by adding a new token -- the \textbf{Graph Context Token} $\text{[GC]}$, which delivers information about the graph context of the document to the BERT. The $\text{[GC]}$ token is placed as the second token in the sequence, following the $\text{[CLS]}$ token and preceding the first-word token. It is composed of two elements: (1) the document node representation and (2) the positional embedding. The sequence of tokens is then normalized.\footnote{We also tested other ways of adding GNN output to the BERT; however, the results were much worse.}

%
%


\subsection{Dataset Selection}\label{sec:dataset}

%
In this work, we focused on solving NLP classification task for text documents that have  interconnections with other documents forming a graph. These connections might be derived from the information about citations, hyperlink references or authorship. 
%

%

We chose Pubmed dataset~\cite{pubmed_dataset}  containing 19,717 scientific articles from the Pubmed database, classified into three topics related to diabetes.\footnote{\url{https://paperswithcode.com/dataset/pubmed}} The dataset forms a graph of abstracts linked through citations. 
%
%
Citation information was obtained from the Pubmed API along with other metadata. An alternative connection method could be through co-authors or shared tags. The articles form a connected, directed, and unweighted graph, with no assigned weights to the citations as they are represented as binary (exist or not) information. A richer representation of citations (calculated by the number of times a given study was referred to in the text) would require access to the full text, which is not provided directly in the Pubmed dataset.

We have found several other interesting datasets that fit into our design; 
%
%
however, due to problems with data accessibility or integrity, it was not possible to use them. For example, in~\cite{hateful_users_on_twitter}, the authors created a curated and manually-labelled dataset containing Twitter data (tweets, users, and their various connections) that were used to investigate hateful user detection. Unfortunately, publicly available data contained preprocessed text without original tweets.
Another example of a dataset that could have been used is a DBpedia dataset~\cite{dbpedia}. It consists of various articles fetched from Wikipedia, each belonging to one of 14 classes, like company, person, etc. However, the original dataset lacked information about hyperlinks between the articles, and it was impossible to get this information from the Wikipedia dump.

\subsection{Experimental Setup}

Due to the limited number of documents in the used Pubmed dataset, overfitting can occur with the BERT architecture. To mitigate this, the training procedure uses smaller batches of 32 documents per epoch, and the final hidden state of the $\text{[CLS]}$ classification token is saved to $T$ matrix for each batch. Backpropagation is performed based on the predictions of the documents from each batch. 
The Adam optimizer~\cite{adam} was used for all training in this study.


The documents are split into three sets: 70\% for training, 10\% for validation, and 20\% for testing. The validation set is used to monitor when the model starts to overfit. The best-performing version is evaluated on the test set.
To ensure unbiased evaluation, we independently split our dataset 10 times into train, validation and test sets using 10 different seed values. Each variant of the models was trained and evaluated on these 10 splits, and the presented results are averaged.

Due to a slight class imbalance in the Pubmed dataset, balanced error and macro $F_1$-score are used as the main evaluation metrics. The balanced error is calculated as 100\% minus balanced accuracy, which is the average accuracy of each class, while the macro $F_1$-score is calculated for each class separately and then averaged over all classes.

\section{Experimental Results and Analysis}\label{sec:results}

In this work, we evaluated GNN+LM models on the Pubmed dataset on the classification task. Before that, basic building blocks (BERT and GNN) were benchmarked independently to establish their default performance on the task.

\begin{table*}[htp]
\centering
\caption{Classification results of all the proposed architectures and their modifications on the Pubmed dataset. Each experiment is repeated 10 times with different dataset splits. Both metrics, error and F1, are balanced, meaning that all classes are treated equally. (Note: (concat/add) - is a strategy for merging representation before a classifier, "skip conn" means that the experiment utilizes a modification of skipping connections.)}
\label{tab_results}
\begin{tabular}{c|r|l|l}
                           & architecture                           & mean error      & mean F1 score    \\ \hline
\multirow{2}{*}{GNN}       & GAT                                    & 22.68\%         & 78.46\% \\
                           & GCN                                    & 14.39\%         & 85.72\% \\ \hline
LM                         & BERT                                   & 8.51\%          & 91.28\%          \\ \hline
\multirow{13}{*}{LM + GNN} & 
                                GNN(BERT)                           & 11.67\%         & 87.84\%          \\
                            & Looped GCBERT                         & 9.00 \%         & 90.79 \%         \\ 
                           & GCBERT + skip conn (concat)            & 8.82\%          & 90.95\%          \\
                           & GCBERT with random $N$ (not trained)   & 8.81\%          & 90.95\%          \\
                           & GCBERT + skip conn (add)               & 8.76\%          & 91.13\%          \\
                           & GCBERT with frozen GNN + skip conn (add)  & 8.66\%          & 91.13\%          \\
                           & Late fusion (add)                         & 8.62\%          & 91.20\%          \\
                           & GCBERT                                    & 8.57\%          & 91.28\%          \\
                           & Looped GCBERT + skip conn (concat)        & 8.49\%          &  91.34 \%        \\
                           & GCBERT with frozen GNN                    & 8.34\%          & 91.50\%          \\
                           & Late fusion (concat)                      & 8.33\%          & 91.43\%          \\
                           & Looped GCBERT + skip conn (add)          & 8.26\%         & 91.61\% \\
                           & GCBERT with frozen GNN + skip conn (concat) & \textbf{7.97\%} & \textbf{91.87\%}
\end{tabular}
\end{table*}

\begin{figure*}[h!]
\centering
\includegraphics[width=\textwidth]{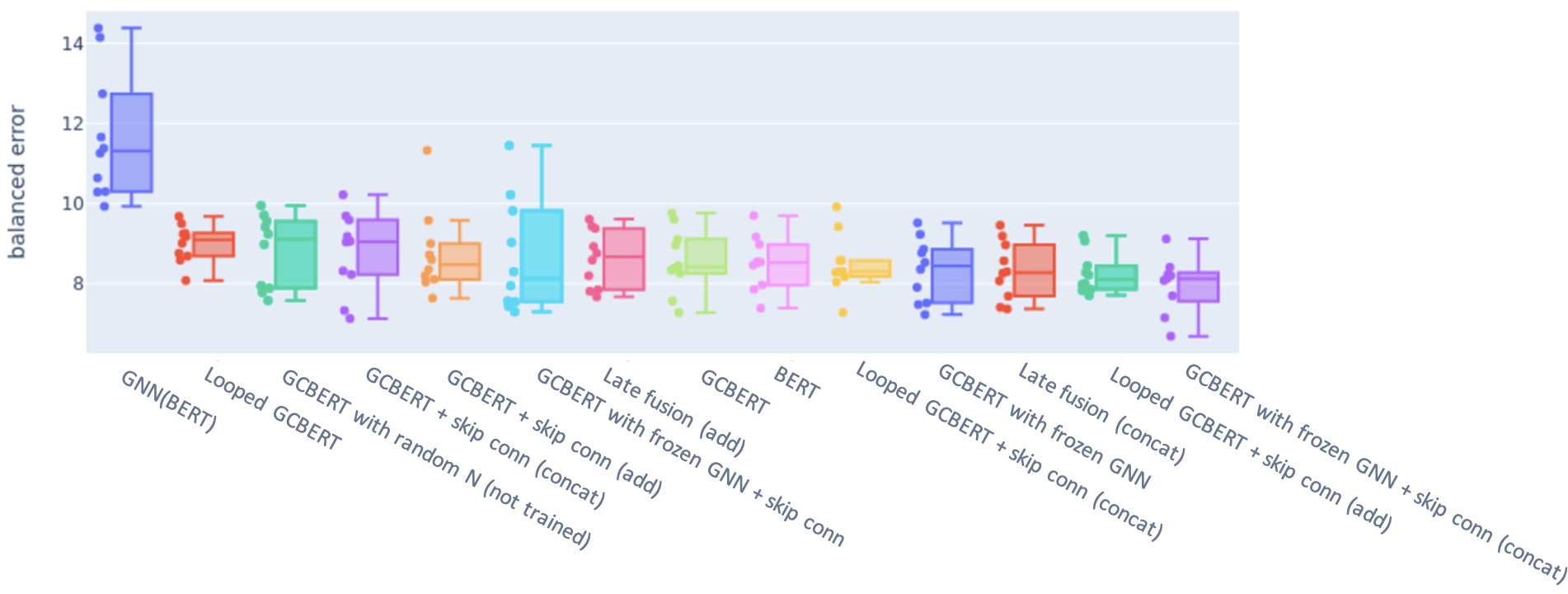}
\caption{Variability of errors in our experiments. Results for each architecture configuration were repeated 10 times with different dataset splits.}
\label{fig_error_variance}
\end{figure*}

We experimented with two benchmark GNNs: Graph Attention Network (GAT) \cite{gat} and Graph Convolutional Network (GCN) \cite{Kipf:2016tc}. When trained alone, these models took a feature matrix $X \in \mathbb{R}^{m \times n}$ ($n$ observations, each of which having dimensionality of $m$) and an adjacency matrix $A \in \mathbb{R}^{n \times n}$ as inputs, where the feature matrix consists of vectorized text representations using TF-IDF. Further experiments were performed with a pre-trained BERT encoder that was fine-tuned on the Pubmed dataset. However, no significant improvements in results have been observed. Therefore, only results obtained using TF-IDF vectorization are reported in Table~\ref{tab_results}. Based on these tests, we can infer that despite supplying GNNs with feature vectors calculated by BERT, they can still not outperform the transformer model, despite having access to additional graph information. There are two potential explanations for this observation: either (1) there is no new discriminative information in the graph that is not already present in the text, or (2) GNNs lose or dilute information from the strong feature vectors produced by BERT when incorporating data from the neighbouring nodes.

BERT architecture alone achieved an average error rate of 8.51\% on the Pubmed test set, significantly lower than all sole GNN models. However, BERT models have a higher number of parameters, making them more flexible than GNNs. Furthermore, in our experiments, the text features extracted for GNNs are based solely on word occurrences, while BERT considers the entire semantic context in the text.

Our evaluation results show that the GNNs achieve diversified performances, with error rates ranging from 22.7\% using GAT to 14.39\% using GCN (see Table~\ref{tab_results} and Figure~\ref{fig_error_variance}).

Table~\ref{tab_results} shows that the compositional architecture  $\text{GNN}(\text{BERT}(x))$ underperforms, with a balanced error rate of 11.67\%. Although it improves the performance of all GNN models, it loses some information learned by BERT, resulting in BERT alone performing better. One possible explanation for this behaviour is that although BERT produces high-quality text representations compared to TF-IDF, GNN is not deep and flexible enough, compared to a dense network, to capture all the information in the input. Moreover, connections between irrelevant documents might lead to confusion in the network. 
Therefore, skip connections were added in later experiments, or the GNN was moved to an earlier stage of the pipeline to overcome this problem.

For \tit{late fusion} architecture, an 8.33\% error rate on the Pubmed test set by concatenating the text and graph representations and 8.62\% by adding them in an element-wise manner is reported. Notably, one of these architectures slightly outperforms the ``plain BERT'' architecture. This was expected, as the model that leverages data from two independent sources should outperform the individual basic models.

The \tit{early fusion} GCBERT architecture, in which GNN is executed before BERT, has variable performance, depending on the modifications used. The basic GCBERT architecture (without any modifications) achieved results comparable to BERT alone, with an error rate of 8.57\%.
However, the addition of skip connections in this architecture can cause the results to be worse. In the case of the concatenation merging strategy, the error rate rose to 8.82\%, and in the case of the addition merging strategy, it rose to 8.76\%. Interestingly, operating on static node representations by finetuning and then freezing the GNN seems to boost the performance of the architecture, with a resulting error rate of 8.34\%.
Additionally, by adding a skip connection with concatenation as a merging strategy, the model reaches a 7.97\% error rate, which significantly outperforms all other architectures, including the baseline BERT.

The cause of the unsatisfactory performance of vanilla GCBERT might result from a new graph token inserted into the input sequence of BERT. In order to assess how this insertion affects the behaviour of the model, an experiment in which BERT was supplied with random vectors generated from Gaussian distribution was conducted. It turns out that the model loses 0.3\% of the accuracy. 
%

Till now, neither architecture allowed an exchange of learned information between components. The \tit{looped GCBERT} architecture addresses this issue by connecting GNN and BERT into one model, components of which can interact with each other. In looped GCBERT, both representations are always dynamic, allowing the network to improve them and adjust to new data constantly. The model reaches 9\% of the error rate. It is a significant improvement over the compositional architecture, which lacked BERT augmentation. 
%
%
However, it is still not enough to outperform basic BERT. The introduction of skip connection with concatenation merge strategy recovers the performance to the level comparable with BERT (8.49\% of error rate), while the one with additive merging strategy outperforms BERT, reaching 8.26\% of the error rate.

In the end, the best architecture tested is the \tit{early fusion GCBERT} with a frozen GNN\footnote{Each time we denoted that GNN is frozen, we trained GNN on our task (adding a classifier for that purpose), and then the GNN was frozen and reused in GCBERT architecture to train the BERT.}, and then added skip connection at the end of the pipeline. It outperforms the baseline by 0.54 percentage points. 

Summarising our experiments, we tried to improve the performance of the BERT model, which has an 8.51\% of error rate. We utilized different modifications and combinations of BERT and GCN architectures to improve the results. The sole GCN model is much weaker in comparison to BERT, having almost twice the higher error rate (14.39\%). 
However, the final results showed that it is worth supporting a strong BERT model with a weaker and much smaller GCN model in order to boost its performance by a decent value.

\section{Concluding Remarks}

Integrating data from different modalities (from various sources or of different structures) can potentially increase the performance of a deep learning model; however, it is still a challenge in deep learning research. 
As we can see from our research, it is not trivial to connect two distinct neural networks that operate on different types of data in such a way as to build a superior architecture compared to its vanilla counterparts. In this study, the goal was to integrate the graph and text data. The neural network was to work on a graph of connected documents instead of a graph of words or topics, as was in the majority of the studies so far.

This study proposed and tested four possible architectural designs: \tit{late fusion}, \tit{early fusion (GCBERT)}, \tit{compositional GNN(BERT)}, and \tit{looped GCBERT}. 
Moreover, we experimented with several modifications that can significantly improve the performance of the models: (1) skip connections, (2) static or dynamic representations, and (3) way of merging text and node representations.
It has been shown that it is possible to outperform the BERT model by supplying it with frozen GNN vectors and adding a skip connection that allows the classifier to analyze both representations. This architecture turned out to be the best and improved the results of BERT by 7\% (relative error). The GCBERT network has only 1.7M parameters, much less than the sole BERT, which has 110M parameters. So, adding a negligibly small GNN component into the BERT model can improve its performance by a decent value.

One can observe that both the general layout and the applied modification have a comparable impact on model performance. 
Adding a skip connection tends to improve the model, but not in all cases. The type of merging of node and text representations also matters significantly. However, there is no clear pattern of which one is better. Its impact depends on other architectural details. 
In some cases, freezing node representations also improves the results, which might seem counterintuitive. 
%
%

In this paper, we experimented with injecting additional graph-based information into language model architecture. The graph information is separate from the semantics of the texts (it is not the classical ontological context). However, it can state additional relevant information sources for the text's understanding. Our results show that there is a place to enrich language models; however, more tests and experiments should be conducted to find the best architectures to capture all the relevant information and fuse them effectively.



Our work is not without limitations. The most crucial is that, on average, we experimented only with one dataset and slightly improved the final results. 
Finding an appropriate dataset was a challenge. We can conclude and indicate as an essential further research direction for the research community to collect and prepare datasets with diversified data sources of various modalities and structures, including graph interconnections of many different types. With such datasets, even artificial ones, our community might extensively experiment with different architectures and ways of fusing data and research the interpretability and contributions of particular modalities and data structures (also of particular neural network components) to the final dataset task.

In further experiments, other tasks should be explored, possibly those in which fine-grained information can have more impact on the final results, e.g. recommendations, topic modelling, information extraction, and machine translation.

Our approach can be easily extended to exchange network components and other language models. Thus it states a very convenient framework for further research and studies on fusing graph-based and text data into neural networks.

\section*{Acknowledgements}
We are particularly grateful to Professor Marcin Paprzycki for his invaluable remarks at the final stage of our work.

This research was carried out with the support of the Laboratory of Bioinformatics and Computational Genomics and the High Performance Computing Center of the Faculty of Mathematics and Information Science Warsaw University of Technology.

Anna Wróblewska wishes to acknowledge that her contribution to this paper was carried out within the framework of Smart Growth Operational Program for 2014-2020, Digital Innovations: grant no POIR.01.01.01-00-0066/22 financed by the National Centre for Research and Development and weSub in Poland.  



\end{document}